\documentclass[fleqn,10pt]{wlpeerj}
\usepackage{url}

\usepackage{color}

\title{TF-CR: Weighting Embeddings for Text Classification}

\author[1]{Arkaitz Zubiaga}
\affil[1]{Queen Mary University of London, UK}
\corrauthor[1]{Arkaitz Zubiaga}{arkaitz@zubiaga.org}

\begin{abstract}
 Text classification, as the task consisting in assigning categories to textual instances, is a very common task in information science. Methods learning distributed representations of words, such as word embeddings, have become popular in recent years as the features to use for text classification tasks. Despite the increasing use of word embeddings for text classification, these are generally used in an unsupervised manner, i.e. information derived from class labels in the training data are not exploited. While word embeddings inherently capture the distributional characteristics of words, and contexts observed around them in a large dataset, they aren't optimised to consider the distributions of words across categories in the classification dataset at hand. To optimise text representations based on word embeddings by incorporating class distributions in the training data, we propose the use of weighting schemes that assign a weight to embeddings of each word based on its saliency in each class. To achieve this, we introduce a novel weighting scheme, Term Frequency-Category Ratio (TF-CR), which can weight high-frequency, category-exclusive words higher when computing word embeddings. Our experiments on 16 classification datasets show the effectiveness of TF-CR, leading to improved performance scores over existing weighting schemes, with a performance gap that increases as the size of the training data grows.
\end{abstract}

\begin{document}

\flushbottom
\maketitle
\thispagestyle{empty}

\section{Introduction}

Text classification is a common task in fields such as natural language processing, information retrieval and information science, consisting in classifying collections of documents into categories \citep{sebastiani2002machine}. Having a collection of documents classified into categories facilitates handling, processing and analysing its underlying content. Automating the process and improving the accuracy of such classification becomes particularly important for large collections or datasets, where manual classification is unmanageable.

The text classification task has evolved substantially since its early days \citep{soergel1985organizing}. While more traditional text classification approaches used vector representations based on the bag-of-words model \citep{boulis2005text}, over the last decade the use of word embeddings \citep{mikolov2013distributed}, or distributed word representations, has gained popularity as a means for vector representation of texts. Word embeddings have become commonplace in tasks such as sentiment analysis \citep{bollegala2016cross}, machine translation \citep{zou2013bilingual}, search \citep{ganguly2015word} or recommender systems \citep{musto2016learning}, as well as across different domains such as biomedicine \citep{chiu2016train} or finance \citep{cortis2017semeval}, outperforming traditional vector representation methods based on a bag-of-words or n-gram model.

Despite the increasing use of word embeddings for text classification in the scientific literature, these are generally fed as pre-trained embeddings that do not consider the distributions of words across categories in the training data. We build on the hypothesis that weighting words according to their prominence in each of the categories in the training data can lead to improved content representation for text classification, leading to improved performance if done properly. To test this, we propose a novel weighting scheme, Term Frequency-Category Ratio (TF-CR), which exploits the category labels from training data to produce an improved vector representation using word embeddings, which is informed by category distributions in the training data. The intuition behind TF-CR is to assign a higher weight to high-frequency, category-exclusive words as observed in the training data.

In this paper, we conduct a set of experiments to assess TF-CR as a weighting scheme for word embeddings in text classification. By experiments with 16 different datasets, we compare the performance of TF-CR as a weighting scheme with the well-known schemes TF-IDF and KLD, with a previously proposed Term Frequency-Term Relevance Ratio (TF-TRR) weighting scheme \citep{ko2012study,ko2015new}, as well as with the standard, unweighted word embeddings. We conduct extensive experimentation evaluating TF-CR with five different embedding models and with different sizes of training data ranging from 1,000 to 90,000 instances. Our experiments show that our proposed scheme, TF-CR, can achieve substantial improvements over all other baseline methods, particularly when the training data increases in size; the baseline method TF-TRR is especially suited to small training sets, whose performance drops as the training data decreases, as opposed to TF-CR. This is, to the best of our knowledge, the first work proposing a weighting scheme that improves the performance of unweighted word embeddings in text classification, extending our preliminary work described in \cite{zubiaga2020exploiting} by adding TF-TRR as an additional baseline, testing on more datasets (16 instead of 8) and performing additional experiments.

\section{Background and Related Work}

Text classification is the task consisting in assigning one (or more) category from a predefined set of categories to each of the elements in a dataset. It is generally tackled as a supervised task, where the classifier is provided with labelled data to build a model which is then applied to unseen test data. The text classification task is applicable for different objectives, including inter alia thematic classification where instances are classified into topics, sentiment analysis where instances are classified into positive, negative or neutral, and spam filtering, where instances are classified as spam or non-spam.

While early methods for text classification in the 1980s were largely manual and rule-based, automated text classification gained ground in the 1990s \citep{sebastiani2002machine}. Text classification models used content representation methods whose vectors had a separate dimension assigned for each word in the dataset. These content representation methods \citep{harris1954distributional} include one-hot encoding, a binary representation indicating if a word is present or not in a document, and the bag-of-words model, where each dimension is assigned a weight equivalent to the frequency of the word in a document. These weights were often altered by using weighting schemes, such as TF-IDF \citep{forman2008bns,salton1988term}. More recently, as new methods such Word2Vec \citep{mikolov2013distributed} and Glove \citep{pennington2014glove} were proposed, word embeddings have become increasingly popular as features to input to text classifiers. Word embeddings can, among others, substantially reduce the vector dimensionality while also capturing semantic properties of words by encoding related words with similar vectors.

Early methods for learning distributed representations of words \citep{bengio2003neural} by leveraging neural probabilistic language models have recently become more popular as embeddings \citep{pilehvar2020embeddings,grohe2020word2vec}. This led to the development of other methods aiming to reduce vector dimensionality by training and modelling word embeddings. While this leads to isolate vectors for each word in a dataset, one of the widely adopted practices for sentence representation is to then get the sum or the average of the word embeddings in the sentence in question \citep{zhang2018aggregating}. Word embeddings have largely replaced bag-of-words representations in a wide variety of tasks \citep{bollegala2016cross,zou2013bilingual,ganguly2015word,musto2016learning} and domains \citep{chiu2016train,cortis2017semeval}.

In this work, we argue that, beyond the inherent characteristics of word embeddings, their effectiveness for text classification can be improved if we consider the distributions of words across classes in the training data. While the use of weighting schemes is common for methods based on one-hot encoding and the bag-of-words model, there is a paucity of research investigating the use of weighting schemes along with word embeddings. A limited body of previous work (e.g. \cite{hu2017user}, \cite{chandu2017tackling}) has applied the TF-IDF weighting scheme on top of word embeddings, but did not explore or propose alternative weighting schemes for word embeddings, which is our objective in this work by proposing TF-CR.

A related line of research is that in which researchers trained separate word embedding models for each category in a dataset by using supervision \citep{tang2014learning,tang2015sentiment,tang2016sentiment,kuang2018class,kuang2019learning}. This is an interesting and complementary line of research, particularly useful for very large datasets with millions of labelled instances, for example in sentiment analysis with datasets created by using distant supervision \citep{go2009twitter}. Our objective here varies in that we aim exploit a weighting scheme over word embedding models trained without supervision, which can be applied jointly with the aforementioned methods but also with datasets which are orders of magnitude smaller.

\section{Weighting Schemes}

Here we discuss the well-known weighting schemes TF-IDF and KLD, as well as the TF-TRR weighting scheme proposed by Ko \cite{ko2012study,ko2015new}. Our work uses all of these schemes as baselines for comparison with our proposed TF-CR scheme.

\subsection{Term Frequency - Inverse Document Frequency (TF-IDF)}

TF-IDF is a well-known weighting scheme originally proposed for information retrieval \cite{jones1972statistical,salton1988term}, which is useful to determine the saliency of a word in a collection. The intuition behind TF-IDF is that words with high frequency occurring in a smaller number of documents in a collection are most helpful for web search. It is the product of two calculations:
\begin{enumerate}
 \item Term Frequency (TF): the number of occurrences of a term, where a higher value increases the overall score of a term.
 \item Inverse Document Frequency (IDF): calculated as the inverse of the document frequency, a high value of IDF indicates that a term occurs in few documents.
\end{enumerate}

TF-IDF is calculated with the following equation:

\begin{equation}
 tf-idf = tf \times idf = f_{t,d} \times \log \frac{|D|}{|\{d \in D: t \in d\}|}
 \label{eq:tfidf}
\end{equation}

where $f_{t,d}$ denotes the frequency of term $t$ in document $d$, $D$ is the entire collection, and $\frac{|D|}{|\{d \in D: t \in d\}|}$ indicates the number of documents that contain $t$.

Different variants are used, for example with logarithmic scaling of TF for reducing the excessive impact of very ihgh TF values.

\subsection{Kullback-Leibler Divergence (KLD)}

KLD \cite{kullback1951information} is a probability metric that allows quantifying the probability of a word occurring in a category in comparison with another category (or another set). For the classification, one can compute the probability of a word in a particular category, compared to the probability of that word in the rest of the categories.
 
The equation for KLD is given in Equation \ref{eq:kld}.

\begin{equation}
 KLD_{w, c} = P(w_c) \log\left(\frac{P(w_c)}{Q(w_r)}\right)
 \label{eq:kld}
\end{equation}

where $P(w_c)$ refers to the probability of the word $w$ in the category $c$, computed as the ratio of words in $c$ which are $w$, and $P(w_r)$ refers to the probability of the word $w$ in the remainder of the categories except $c$, computed as the ratio of words in the remained of the categories which are $w$.

\subsection{TF-TRR}

TF-TRR is a weighting scheme proposed by \cite{ko2012study,ko2015new}. TF-TRR was proposed as variant of TF-IDF, where IDF is replaced with TRR. TRR is a simple metric that divides the probability of a word $w$ in category $c$, divided by the probability of the word $w$ in the rest of the categories, as shown in Equation \ref{eq:trr}.

\begin{equation}
 TR-TRR = (\log tf_w + 1) \log\left(\frac{P(w|c)}{P(w|r)} + \alpha\right)
 \label{eq:trr}
\end{equation}

where P(w|c) is the probability of the word $w$ in the category $c$, and P(w|r) is the probability of the word $r$ in the remainder of the categories. Alpha is a constant to avoid negative values resulting from the logarithm. We set this value to 1.2 for our experiments.

\section{The TF-CR Weighting Scheme}
\label{sec:tfcr}

We propose a novel weighting scheme named Term Frequency-Category Ratio (TF-CR). TF-CR is a simple weighting scheme that computes the product of the importance of a word within a category (Term Frequency, TF) and the distribution of the word across all categories (Category Ratio, CR). Both TF and CR are computed for each word $w$ within each category $c$:

\begin{itemize}
 \item TF is calculated as the ratio of words in a category that are $w$, i.e. $TF_{wc} = \frac{|w_c|}{N_c}$, where $|w_c|$ is the number of occurrences of $w$ in $c$, and $N_c$ is the total number of word occurrences in $c$.
 \item CR is calculated as the ratio of occurrences of $w$ that occur within the category $c$, i.e. $CR_{wc} = \frac{|w_c|}{|w|}$, where $|w|$ denotes the number of occurrences of $w$ across all categories.
\end{itemize}

The final TF-CR score for word $w$ in category $c$ is the product of both metrics (see Equation \ref{eq:tfcr}).

\begin{equation}
 TF-CR_{w, c} = \frac{|w_c|}{N_c} * \frac{|w_c|}{|w|} = \frac{|w_c|^2}{N_c * |w|}
 \label{eq:tfcr}
\end{equation}

The intuition behind TF-CR is to assign a high weight to words that occur exclusively and with high frequency within a category. Low-frequency words or words that occur across all categories will get lower scores.

\section{Experiments}

\subsection{Applying Weighting Schemes on Embeddings}
\label{ssec:applying}

The weighting schemes provide a weight for each word-category pair. Once we have these weights, we need to come up with a way of representing documents in our datasets using word embeddings in combination with these weights. Using category-specific weights for text classification presents the challenge that the category is unknown for the test data. While one knows the weights that best apply to training data based on their labelled category, this is not applicable to test data. \cite{ko2015new} proposed to rely on the label for weighting training data, whereas for the test data they chose the category for which the aggregate of weights would be higher. While this seems reasonable in their case using bag-of-words representations, due to the large dimensionality, we propose another alternative given that word embeddings are much lower in dimensionality. Instead of having to choose a category, we concatenate the weighted representations for all $N$ categories, both in the case of training and test data. Therefore, if the word embedding model uses $d$ dimensions, our weighted representation contains $N \times d$ dimensions.

\subsection{Datasets}

We use 16 datasets pertaining to very different classification tasks ranging from sentiment analysis to thematic classification:

\begin{enumerate}
 \item \textbf{RepLab polarity dataset \citep{replab2013overview}:} A dataset of 84,745 tweets mentioning companies, annotated for polarity as positive, negative or neutral.\footnote{\url{http://nlp.uned.es/replab2013/}} While the original dataset contains more than 140,000 tweets, we use the 84,745 tweets we were able to retrieve, due to other tweets being deleted \cite{zubiaga2018longitudinal}.
 \item \textbf{ODPtweets \citep{zubiaga2013harnessing}:} a large-scale dataset with nearly 25 million tweets, each categorised into one of the 17 categories of the Open Directory Project (ODP).
 \item \textbf{Restaurant reviews \cite{jiang2019leveraging}:} a large dataset of 14,542,460 TripAdvisor restaurant reviews with their associated star rating ranging from 1 to 5.
 \item \textbf{Attraction reviews \cite{jiang2019leveraging}:} 6,358,253 TripAdvisor attraction reviews with star ratings ranging from 1 to 5.
 \item \textbf{Hotel reviews \cite{jiang2019leveraging}:} 3,598,292 TripAdvisor hotel reviews with star ratings ranging from 1 to 5.
 \item \textbf{Book reviews \cite{mcauley2015image,he2016ups}:} from a large Amazon review dataset released by McAuley et al., we extract the set of 22,507,155 reviews pertaining to books. Each book review includes a star rating from 1 to 5.
 \item \textbf{Clothing reviews \cite{mcauley2015image,he2016ups}:} from the same Amazon review dataset, this includes the set of 5,748,920 reviews pertaining to clothing products, with their star ratings from 1 to 5.
 \item \textbf{Homeware reviews \cite{mcauley2015image,he2016ups}:} from the same Amazon review dataset, this includes the set of 4,253,926 reviews pertaining to homeware, with their star ratings from 1 to 5.
 \item \textbf{SemEval sentiment tweets \citep{rosenthal2017semeval}:} we aggregate all annotated tweets from the SemEval Twitter sentiment analysis task from 2013 to 2017. The resulting dataset contains 61,767 tweets.
 \item \textbf{Distantly supervised sentiment tweets:} by using a large collection of tweets from January 2013 to September 2019 released on the Internet Archive\footnote{\url{https://archive.org/details/twitterstream}}, we produce a dataset of tweets annotated for sentiment analysis by using distant supervision following \citep{go2009twitter}, leading to tweets annotated as positive or negative. The resulting dataset contains 33,203,834 tweets.\footnote{\url{http://www.zubiaga.org/datasets/sentiment1319/}}
 \item \textbf{Hate speech dataset \citep{founta2018large}:} a dataset of 99,996 tweets, each categorised into one of \{abusive, hateful, spam, normal\}.
 \item \textbf{Newsspace200 \citep{del2005ranking}:} a dataset of nearly 500K news articles, each categorised into one of 14 categories, including business, sports entertainment.\url{http://groups.di.unipi.it/~gulli/AG_corpus_of_news_articles.html}
 \item \textbf{20 Newsgroups:} a collection of nearly 20,000 newsgroup documents, pertaining to 20 different newsgroups, which are used as categories.\footnote{\url{http://qwone.com/~jason/20Newsgroups/}}
 \item \textbf{DBPedia \citep{zhang2015text}:} a dataset built from DBPedia's curated ontology, which contains 630,000 instances distributed across 14 categories.
 \item \textbf{Sogou news \citep{zhang2015text}:} a collection of 510,000 news articles across 5 categories. This dataset contains documents in Chinese languages, as opposed to the rest of the datasets in English.
 \item \textbf{Yahoo! Answers \citep{zhang2015text}:} a collection of 1,460,000 questions and answers from Yahoo!, categorised across 10 classes.
\end{enumerate}

For all datasets, we randomly sample 100,000 instances, except for those with fewer instances. These samples are used to conduct our experiments.

\subsection{Word Embedding Models \& Classifiers}
\label{ssec:embedding}

We perform experiments with a wide range of word embedding models and classifiers to test the robustness of the TF-CR weighting scheme in different settings.

We use five different word embedding models:

\begin{itemize}
 \item \textbf{w2v:} an in-domain Word2Vec model trained from the dataset in question, i.e. resulting in 16 embedding models that we train. The entire, original dataset is used in each case, not only the smaller sample used for classification.
 \item \textbf{gw2v:} Google's Word2Vec model.
 \item \textbf{tw2v:} a Twitter Word2Vec model\footnote{\url{https://fredericgodin.com/software/}} \cite{godin2015multimedia}.
 \item \textbf{cglove:} GloVe embeddings trained from Common Crawl.
 \item \textbf{wglove:} GloVe embeddings trained from Wikipedia.\footnote{\url{https://nlp.stanford.edu/projects/glove/}}
\end{itemize}

We use two different classifiers for these experiments, SVM and Logistic Regression, which are known to perform reasonably well and efficiently in a way that allow us to perform experiments at scale.

\section{Results}

This section shows and discusses the results of our experiments, focusing on the comparison of three different dimensions revolving around the comparison of weighting schemes: classifiers, embedding models and training data sizes. We report macro-F1 values as performance scores. All performance scores reported are the result of averaging 10-fold cross-validation experiments.

\subsection{Performance Across Classifiers}

\begin{table}[htp]
 \centering
 \begin{tabular}{| l || r | r | r | r | r || r | r | r | r | r |}
  \hline
  & \multicolumn{5}{c||}{SVM} & \multicolumn{5}{c|}{Logistic Regression} \\
  \cline{2-11}
  \multicolumn{1}{|c||}{dataset} & \multicolumn{1}{c|}{$\emptyset$} & \multicolumn{1}{c|}{tfidf} & \multicolumn{1}{c|}{kld} & \multicolumn{1}{c|}{tftrr} & \multicolumn{1}{c||}{tfcr} & \multicolumn{1}{c|}{$\emptyset$} & \multicolumn{1}{c|}{tfidf} & \multicolumn{1}{c|}{kld} & \multicolumn{1}{c|}{tftrr} & \multicolumn{1}{c|}{tfcr} \\
  \hline
  \hline
  20ng & .667 & .876 & .862 & .876 & \textbf{.912} & .607 & .822 & .826 & .740 & \textbf{.896} \\
  dbpedia & .959 & .961 & .964 & .940 & \textbf{.980} & .937 & .963 & .952 & .929 & \textbf{.978} \\
  hate & .591 & .522 & .627 & .615 & \textbf{.633} & .581 & .502 & .623 & .587 & \textbf{.633} \\
  ns200 & .557 & .528 & .587 & .561 & \textbf{.592} & .560 & .552 & .590 & .501 & \textbf{.608} \\
  odp & .369 & .353 & .365 & .436 & \textbf{.447} & .376 & .375 & .367 & .343 & \textbf{.453} \\
  replab & .407 & .375 & .382 & .400 & \textbf{.408} & .403 & .366 & .377 & .363 & \textbf{.407} \\
  rev-attr & .381 & .327 & .367 & .375 & \textbf{.391} & .379 & .349 & .369 & .312 & \textbf{.408} \\
  rev-book & .443 & .348 & .435 & .433 & \textbf{.455} & .440 & .369 & .432 & .422 & \textbf{.468} \\
  rev-clot & .476 & .389 & .460 & .462 & \textbf{.481} & .471 & .404 & .453 & .393 & \textbf{.494} \\
  rev-home & .476 & .383 & .462 & .474 & \textbf{.485} & .465 & .394 & .454 & .410 & \textbf{.495} \\
  rev-hote & .531 & .431 & .520 & .501 & \textbf{.539} & .522 & .453 & .518 & .412 & \textbf{.549} \\
  rev-rest & .520 & .432 & .504 & .504 & \textbf{.522} & .514 & .453 & .505 & .425 & \textbf{.537} \\
  semeval & .610 & .519 & .565 & .600 & \textbf{.630} & .602 & .521 & .557 & .592 & \textbf{.629} \\
  sentiment & .728 & .630 & .664 & .707 & \textbf{.747} & .727 & .630 & .665 & .714 & \textbf{.748} \\
  sogou & .910 & .891 & .921 & .886 & \textbf{.925} & .886 & .883 & .907 & .894 & \textbf{.910} \\
  yahoo & .686 & .636 & .642 & .630 & \textbf{.687} & .674 & .660 & .633 & .539 & \textbf{.693} \\
  \hline
 \end{tabular}
 \caption{Macro-average F1 scores for different weighting schemes across the 16 datasets, using SVM and logistic regression as classifiers.}
 \label{tab:classifiers}
\end{table}

Table \ref{tab:classifiers} shows results for the experiments on 16 datasets using the different weighting schemes with SVM and Logistic Regression as classifiers. These experiments use in-domain word embedding models that we pre-trained (w2v), i.e. using the larger dataset available in each case. Note however that the pre-training of word embedding models is unsupervised, i.e. not using any labels. We can observe that, in the case of both classifiers, these results show a consistent improvement of TF-CR over all the baseline methods. These experiments show all the available training instances for each dataset (i.e. up to 90,000), showing that in these settings TF-CR leads to substantial improvements over all other methods. Results for the other weighting schemes, TF-IDF, KLD and TF-TRR, are largely inconsistent across datasets, with no clear winner between them, and not always outperforming the unweighted baseline.

Given similar trends in scores, we focus on the logistic regression classifier for subsequent experiments, in the interest of brevity and clarity.

\subsection{Performance Across Embedding Models}

\begin{table}[htp]
 \centering
 \begin{tabular}{| l || r | r | r | r | r || r | r | r | r | r |}
  \hline
  & \multicolumn{5}{c||}{gw2v} & \multicolumn{5}{c|}{tw2v} \\
  \cline{2-11}
  \multicolumn{1}{|c||}{dataset} & \multicolumn{1}{c|}{$\emptyset$} & \multicolumn{1}{c|}{tfidf} & \multicolumn{1}{c|}{kld} & \multicolumn{1}{c|}{tftrr} & \multicolumn{1}{c||}{tfcr} & \multicolumn{1}{c|}{$\emptyset$} & \multicolumn{1}{c|}{tfidf} & \multicolumn{1}{c|}{kld} & \multicolumn{1}{c|}{tftrr} & \multicolumn{1}{c|}{tfcr} \\
  \hline
  \hline
  20ng & .701 & .884 & .843 & .784 & \textbf{.922} & .705 & .893 & .860 & .794 & \textbf{.930} \\
  dbpedia & .961 & .965 & .961 & .942 & \textbf{.980} & .959 & .962 & .964 & .938 & \textbf{.980} \\
  hate & \textbf{.655} & .550 & .642 & .630 & .654 & \textbf{.661} & .556 & .643 & .638 & .648 \\
  ns200 & .534 & .508 & .572 & .518 & \textbf{.599} & .544 & .507 & .586 & .514 & \textbf{.595} \\
  odp & .310 & .341 & .322 & .301 & \textbf{.436} & .325 & .354 & .362 & .311 & \textbf{.458} \\
  replab & .409 & .357 & .354 & .380 & \textbf{.436} & .422 & .379 & .364 & .405 & \textbf{.444} \\
  rev-attr & .367 & .298 & .378 & .325 & \textbf{.394} & .374 & .298 & .386 & .340 & \textbf{.394} \\
  rev-book & .423 & .307 & .441 & .408 & \textbf{.447} & .429 & .304 & .453 & .417 & \textbf{.452} \\
  rev-clot & .439 & .345 & .461 & .379 & \textbf{.481} & .460 & .356 & .472 & .407 & \textbf{.483} \\
  rev-home & .430 & .340 & .465 & .391 & \textbf{.476} & .454 & .345 & .475 & .411 & \textbf{.482} \\
  rev-hote & .517 & .386 & .526 & .422 & \textbf{.537} & .525 & .386 & .532 & .432 & \textbf{.533} \\
  rev-rest & .497 & .388 & .508 & .439 & \textbf{.525} & .503 & .390 & .516 & .441 & \textbf{.522} \\
  semeval & .615 & .515 & .564 & .595 & \textbf{.638} & .635 & .532 & .577 & .611 & \textbf{.638} \\
  sentiment & .680 & .601 & .649 & .706 & \textbf{.726} & .721 & .647 & .663 & .726 & \textbf{.748} \\
  sogou & .634 & .617 & .686 & .588 & \textbf{.724} & .616 & .616 & .686 & .580 & \textbf{.786} \\
  yahoo & .672 & .625 & .643 & .587 & \textbf{.696} & .670 & .613 & .645 & .578 & \textbf{.687} \\
  \hline
  \hline
  & \multicolumn{5}{c||}{cglove} & \multicolumn{5}{c|}{wglove} \\
  \cline{2-11}
  \multicolumn{1}{|c||}{dataset} & \multicolumn{1}{c|}{$\emptyset$} & \multicolumn{1}{c|}{tfidf} & \multicolumn{1}{c|}{kld} & \multicolumn{1}{c|}{tftrr} & \multicolumn{1}{c||}{tfcr} & \multicolumn{1}{c|}{$\emptyset$} & \multicolumn{1}{c|}{tfidf} & \multicolumn{1}{c|}{kld} & \multicolumn{1}{c|}{tftrr} & \multicolumn{1}{c|}{tfcr} \\
  \hline
  \hline
  20ng & .766 & .899 & .881 & .829 & \textbf{.928} & .748 & .893 & .880 & .807 & \textbf{.928} \\
  dbpedia & .965 & .957 & .966 & .933 & \textbf{.978} & .966 & .956 & .967 & .928 & \textbf{.978} \\
  hate & \textbf{.656} & .556 & .642 & .630 & .639 & \textbf{.645} & .544 & .640 & .637 & .644 \\
  ns200 & .573 & .517 & \textbf{.596} & .525 & .592 & .560 & .514 & \textbf{.590} & .443 & .586 \\
  odp & .342 & .355 & .375 & .327 & \textbf{.451} & .296 & .315 & .346 & .269 & \textbf{.426} \\
  replab & .432 & .384 & .370 & .383 & \textbf{.448} & .421 & .375 & .395 & .400 & \textbf{.443} \\
  rev-attr & .368 & .295 & \textbf{.388} & .321 & .376 & .351 & .286 & \textbf{.382} & .314 & .375 \\
  rev-book & .424 & .307 & \textbf{.449} & .407 & .419 & .408 & .296 & \textbf{.444} & .402 & .423 \\
  rev-clot & .450 & .347 & \textbf{.469} & .382 & .462 & .429 & .335 & \textbf{.467} & .378 & .462 \\
  rev-home & .441 & .341 & \textbf{.476} & .390 & .462 & .419 & .329 & \textbf{.472} & .388 & .461 \\
  rev-hote & .519 & .388 & \textbf{.532} & .413 & .504 & .504 & .377 & \textbf{.531} & .409 & .504 \\
  rev-rest & .497 & .387 & \textbf{.514} & .428 & .499 & .476 & .379 & \textbf{.511} & .427 & .498 \\
  semeval & .624 & .521 & .575 & .601 & \textbf{.631} & .597 & .504 & .566 & .580 & \textbf{.627} \\
  sentiment & .692 & .626 & .653 & .714 & \textbf{.742} & .672 & .592 & .646 & .702 & \textbf{.731} \\
  sogou & .758 & .755 & .862 & .750 & \textbf{.895} & .696 & .691 & .779 & .647 & \textbf{.835} \\
  yahoo & \textbf{.689} & .613 & .651 & .560 & .680 & .667 & .615 & .645 & .543 & \textbf{.682} \\
  \hline
 \end{tabular}
 \caption{Macro-average F1 scores for different weighting schemes across the 16 datasets, using four different pre-trained word embedding models with a logistic regression classifier.}
 \label{tab:schemes}
\end{table}

In addition to the previous experiments using in-domain word embedding models (w2v), here we test four other embedding models, as described in Section \ref{ssec:embedding}: gw2v, tw2v, cglove and wglove. These results large reaffirm the results with the in-domain embeddings showing best-performing results for tf-cr, with the following exceptions:
\begin{itemize}
 \item the \textit{hate} dataset, where unweighted embeddings outperform tf-cr.
 \item the glove embeddings, where kld outperforms tf-cr on several datasets, and the unweighted approach outperforms tf-cr for the \textit{yahoo} dataset.
\end{itemize}

In the rest of the cases, tf-cr is still the best-performing method, regardless of the word embedding model used.

\subsection{Performance by Training Data Size}

In previous experiments we have used all of the up to 90,000 training instances available for each dataset. Here we perform additional experiments using varying numbers of training instances, ranging from 1,000 to 90,000. This allows us to assess the extent to which weighting schemes can help with varying sizes of training data, provided that calculations of weights using these schemes are done solely from the training data available in each case. We randomly sample training instances in each training scenario, keeping the random sample consistent across different experiments with the same training size, and incrementally adding instances, i.e. a training set with 9,000 instances contains all of the training instances of that with 8,000 plus an additional 1,000 instances.

\begin{figure}[htbp]
 \centering
 \includegraphics[width=\linewidth]{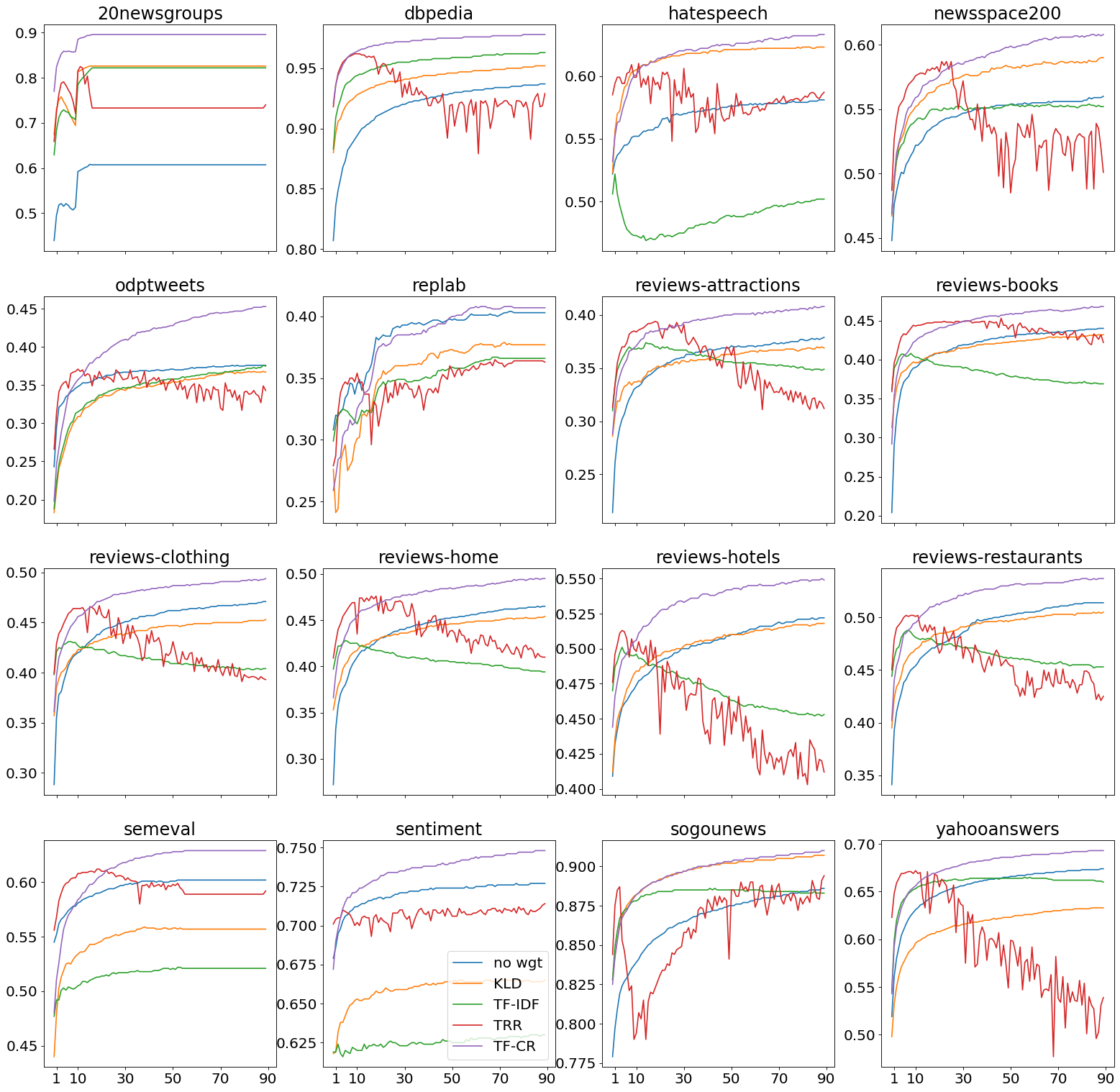}
 \caption{Macro-average F1 scores for varying sizes of training data across the 16 datasets using an in-domain embedding model. Performance scores plotted with a step size of 1,000 instances, ranging from 1,000 to 90,000.}
 \label{fig:results-size}
\end{figure}

Figure \ref{fig:results-size} shows results for the different weighting methods across the 16 datasets as the training data size increases. They again all show a similar tendency:

\begin{itemize}
 \item The baseline method trr proves to perform very well and generally the best when the training size is small.
 \item tf-cr overcomes trr as the training size increases, performing best for mid- to large-sized training sets in all cases.
 \item The performance of tf-cr increases as the training size increases. This proves that, contrary to trr, tf-cr benefits from having more observations in the training data, making the most of label and word distributions in the data when the training size increases.
\end{itemize}

\section{Conclusions}

In this work we have introduced Term Frequency-Category Ratio (TF-CR), a weighting scheme that leverages word distributions across classes in the training data to weight word embedding representations. Different from the TF-IDF weighting scheme, which was designed for information retrieval, the TF-CR weighting scheme is designed for text classification. The intuition behind TF-CR is to assign high weights to high-frequency words that predominantly occur within a particular category, therefore indicating that the word is highly relevant to the category.

We compare the performance of the proposed TF-CR scheme with four other baseline methods: TF-IDF, KLD, TF-TRR and an unweighted baseline. Experimenting on 16 datasets, we observe consistent results showing that TF-CR outperforms all other baseline methods when we have medium to large sized training sets at hand. While the previously proposed method, TF-TRR, performs best for small training sets, its performance drops as training data increases, where TF-CR achieves a significant gain over the other methods.

The code implementing TF-CR can be found at \url{https://github.com/azubiaga/tfcr}.

\section*{Acknowledgments}

This research utilised Queen Mary's Apocrita HPC facility, supported by QMUL Research-IT. \url{http://doi.org/10.5281/zenodo.438045}

\bibliography{tfcr}

\end{document}